\documentclass[conference]{IEEEtran}
\IEEEoverridecommandlockouts
\usepackage{cite}
\usepackage{amsmath,amssymb,amsfonts}
\usepackage{algorithmic}
\usepackage{graphicx}
\usepackage{textcomp}
\usepackage{xcolor}
\usepackage{booktabs} 
\usepackage{siunitx}
\usepackage{caption,subcaption}

\def\BibTeX{{\rm B\kern-.05em{\sc i\kern-.025em b}\kern-.08em
    T\kern-.1667em\lower.7ex\hbox{E}\kern-.125emX}}
\begin{document}

\title{ Testing the Efficacy of Hyperparameter Optimization Algorithms in Short-Term Load Forecasting\\

}

\author{\IEEEauthorblockN{1\textsuperscript{st} Tuğrul Cabir Hakyemez}
\IEEEauthorblockA{\textit{Dept. of Management Information Systems} \\
\textit{Istanbul Bilgi University}\\
Istanbul, Turkey \\
tugrul.hakyemez@bilgi.edu.tr}
\and
\IEEEauthorblockN{2\textsuperscript{nd}Omer Adar}
\IEEEauthorblockA{\textit{Samsun Atakum Bahcesehir Anadolu Lisesi}\\
Samsun, Turkey \\
omeradar155@gmail.com}
}

\maketitle

\begin{abstract}
Accurate forecasting of electrical demand is essential for maintaining a stable and reliable power grid, optimizing the allocation of energy resources, and promoting efficient energy consumption practices. This study investigates the effectiveness of five hyperparameter optimization (HPO) algorithms—Random Search, Covariance Matrix Adaptation Evolution Strategy (CMA-ES), Bayesian Optimization, Partial Swarm Optimization (PSO), and Nevergrad Optimizer (NGOpt) across  univariate and multivariate Short-Term Load Forecasting (STLF) tasks. Using the Panama Electricity dataset (n=48,049), we evaluate HPO algorithms' performances on a surrogate forecasting algorithm, XGBoost, in terms of accuracy (i.e., MAPE, R2) and runtime. Performance plots visualize these metrics across varying sample sizes from 1,000 to 20,000, and Kruskal-Wallis tests assess the statistical significance of the performance differences. Results reveal significant runtime advantages for HPO algorithms over Random Search. In univariate models, Bayesian optimization exhibited the lowest accuracy among the tested methods. This study provides valuable insights for optimizing XGBoost in the STLF context and identifies areas for future research.
\end{abstract}

\begin{IEEEkeywords}
Time Series Forecasting, Short Term Load Forecasting, Hyperparameter Optimization, XGBoost, Kruskal- Wallis test
\end{IEEEkeywords}

\section{Introduction}
Accurate load forecasting (LF) is crucial for ensuring a reliable power supply and efficient energy resource management [1]. It involves predicting future energy demand using electricity usage data at various aggregation levels, which reveals consumption patterns over time. As a subfield of LF, short-term load forecasting (STLF) focuses on predictions ranging from minutes to days ahead, while long-term load forecasting (LTLF) extends up to 20 years into the future [2]. However, achieving accurate LF is challenging due to factors including household routines, appliance usage, holidays, seasons, weather conditions [3].  Fortunately, a wide range of methods, including traditional statistical approaches (e.g., ARIMA, Linear Regression) [4] and machine learning algorithms (e.g., XGBoost, LSTM, ANN) [5,6], can incorporate usage data and relevant features to make accurate forecasts.  However, the performance of these methods depends heavily on the optimal configuration of hyperparameters

Hyperparameter optimization (HPO) is a crucial step in machine learning, as the values of these non-trainable parameters [7]  can significantly impact model performance [8]. The optimization process seeks the optimal values for key model parameters such as learning rate, number of hidden layers in neural networks, or number of trees in ensemble tree-based methods like XGBoost. The primary tradeoff lies in balancing the search space: a broader space increases the chance of finding optimal values but at the cost of higher runtime and computational complexity. Traditional algorithms like grid and random search are less sensitive to this balance.  In response, researchers have developed "intelligent" optimization techniques (e.g., Bayesian, evolutionary) that can conduct an informed search by narrowing down the parameter space under the guidance of a performance criterion, usually a loss metric [9]. Given the challenges of making accurate short-term load forecasts, optimizing hyperparameters is particularly crucial in this domain. HPO not only increases accuracy rates but also shortens the convergence time of algorithms used in STLF [10,11]. Previous research has explored suitable HPO algorithms for STLF [4, 6], but questions remain regarding the statistical significance of performance differences, scalability across sample sizes, and robustness across datasets at different aggregation levels. 

Previous research has effectively utilized HPOs for both classification [12, 13,14] and regression tasks [5,6,15]. In the context of STLF, various HPO algorithms have been employed to optimize the performance of specific forecasting models [16]. However, a comprehensive comparison of these HPO algorithms across different datasets and sample sizes, along with rigorous statistical analysis, is lacking in the existing literature. Table 1 summarizes key research in HPO for STLF and highlights the unique contributions of the current study

In this study, we compare five state-of-the-art HPO algorithms—random search,\\ Covariance Matrix Adaptation Evolution Strategy (CMA-ES), Bayesian optimization, Partial Swarm Optimization (PSO), and Nevergrad Optimizer (NGOpt)—in both univariate and multivariate data configurations. We evaluate their performance using accuracy metrics (MAPE,$R^2$) and runtime metrics in the context of STLF. We used samples of varying sizes from a publicly available dataset: Panama National Electricity Demand [17]. We employed XGBoost due to its proven effectiveness in load forecasting tasks and its sensitivity to hyperparameter tuning [18,19]. The performances of the HPO algorithms are displayed on algorithm comparison plots that show the values of individual metrics for increasing sample sizes across model configurations. Lastly, a Kruskal-Wallis test examines the statistical significance of performance differences between the HPO algorithms optimizing the performance of XGBoost.
The contributions of this study can be summarized as follows:
\begin{itemize}
\item We systematically evaluate a wide range of state-of-the-art HPO algorithms, including NGOpt, a meta-algorithm that adapts based on the task, in the context of short-term load forecasting (STLF).
\item We visualize the performance of HPO algorithms across increasing sample sizes using performance plots, providing insights into their scalability as data volume grows.
\item We propose the use of a non-parametric Kruskal-Wallis test to assess the statistical significance of performance differences between HPO algorithms, ensuring robust comparisons.
\end{itemize}

The remainder of this paper is organized as follows. Section 2 details  our research methodology. Section 3 discusses the results of our experiments, and Section 4 offers concluding remarks and directions for future research.

\begin{table*}
\centering
\caption{Key Research}
\label{tab:algorithm_comparison}
\begin{tabular}{|p{1.5cm}|p{4.25cm}|p{3.5cm}|p{3.5cm}|p{1.8cm}|p{2.2cm}|}  
\hline
\raggedright
\textbf{Study} & \textbf{Dataset} & \textbf{ML Algorithms} & \textbf{HPO Algorithms} & \textbf{Prediction Horizon} & \textbf{Performance Metrics} \\ 
\hline
[5] & Solar energy output data (Brisbane, Australia) & ARIMA, XGBoost, MLP, RBF & PSO & Day, week, month & NRMSE \\ 
\hline
[15] & Public Irish CER dataset & Online SVR, MLP, DLNet, Bagging, XRT, RF, XGBoost & PSO, Cuckoo optimization algorithm & 30 min, 1-hour, day & MAPE \\ 
\hline
[6] & Paraguay National Electrical demand dataset & LSTM & Grid Search, Random Search, Genetic Algorithm, TPE & Day & RMSE, MAPE, R\textsuperscript{2} \\ 
\hline
[16] & Johor City Hourly Electrical Demand & LGBM, XGBoost, MLP & PSO, Bayesian, Random Search, Evolutionary strategy, SA & Day & MAE, MAPE, RMSE, MSLE, MAE, R\textsuperscript{2}, Runtime \\ 
\hline
[4] & Johor City Hourly Electrical Demand & Linear Regression, DT, SVM, Ensemble, GPR, Neural Networks & Grid search, random search, Bayesian optimization & 1-hour & RMSE, MAPE, MAE, MSE, R\textsuperscript{2}, Training time \\ 
\hline
\textbf{This study} & \textbf{Panama National Electricity demand} & \textbf{XGBoost} & \textbf{Random Search, CMA-ES, Bayesian, PSO, NGOpt} & \textbf{1-hour} & \textbf{MAPE, R\textsuperscript{2}, Runtime} \\ 
\hline
\end{tabular}
\end{table*}

\section{Methodology}

\subsection{Problem Definition}\label{AA}
The STLF problem involves predicting electricity demand (Y) at future hourly intervals. This prediction is based on historical hourly demand data, formulating the problem as a time series prediction task. The specific goal is to predict the demand at the next hour, Y(t+1), given a sequence of past demand observations.
For univariate prediction, the model leverages a time series of historical demand values, denoted as Y(t-S):t, where S represents the number of previous hours considered as input features.
\begin{equation}
[Y(t-S):t] \rightarrow f\,Y(t+1) 
\end{equation}
For multivariate prediction, the model incorporates additional features, such as weather variables (e.g., temperature, relative humidity, wind speed) and categorical indicators (e.g., school days and holidays), represented by the feature matrix X(t-S):t. This matrix has dimensions S x D, where S denotes the number of previous hours considered, and D represents the total number of input features.
\begin{equation}
[Y(t-S):t, X(t-S):t] \rightarrow f\,Y(t+1)
\end{equation}

In summary, the objective is to predict the next hour's electricity demand at country and household levels using historical demand data and potentially other relevant features such as weather conditions or calendar information.

\subsection{Datasets}
This study utilizes an illustrative sample from the Panama national electricity load dataset [17]. The dataset comprises hourly Panama national electricity demand measurements between January 2015 and June 2020 (n=48,049), with additional continuous features (i.e., temperature, humidity, wind speed, precipitation) and categorical features (i.e., holiday, school). Within this period, there are 3,025 holidays and 34,970 school days. Table 2 displays key descriptive statistics for the continuous variables in both datasets. The Panama national demand dataset is complete, with no missing values. To avoid duplicate observations, we first removed any rows with identical indices. Next, we applied min-max scaling to this dataset to normalize the features and address potential unit and range discrepancies between columns. 

\begin{table}
\centering
\caption{Descriptive Statistics of Continuous Variables by Location}
\label{tab:descriptives}
\begin{tabular}{|p{1.5cm}|p{1.5cm}|p{1.1cm}|p{1.1cm}|p{1.1cm}|p{0.7cm}|}
\hline
\textbf{Variable} & \textbf{Description} & \textbf{Min} & \textbf{Max} & \textbf{Mean} & \textbf{SD.} \\ 
\hline
Consumption & National electricity load (MWh) & 85.19 & 1754.88 & 1182.86 & 192.06 \\ 
\hline
T2M\_toc & Temperature (\textcelsius) & 22.95 & 35.03 & 27.39 & 1.67 \\ 
\hline
QV2M\_toc & Relative Humidity & 0.012 & 0.02 & 0.01 & 0.001 \\ 
\hline
W2M\_toc & Wind Speed (m/s) & 0 & 39.22 & 13.39 & 7.29 \\ 
\hline
TQL\_toc & Liquid precipitation (liters/m\textsuperscript{2}) & 0 & 0.52 & 0.079 & 0.06 \\ 
\hline
T2M\_san & Temperature (\textcelsius) & 19.76 & 39.06 & 26.92 & 3.01 \\ 
\hline
QV2M\_san & Relative Humidity & 0.01 & 0.02 & 0.01 & 0.001 \\ 
\hline
W2M\_san & Wind Speed (m/s) & 0.06 & 24.48 & 7.04 & 4.01 \\ 
\hline
TQL\_san & Liquid precipitation (liters/m\textsuperscript{2}) & 0 & 0.48 & 0.1 & 0.08 \\ 
\hline
T2M\_dav & Temperature (\textcelsius) & 19.93 & 34.21 & 24.71 & 2.41 \\ 
\hline
QV2M\_dav & Relative Humidity & 0.009 & 0.021 & 0.01 & 0.001 \\ 
\hline
W2M\_dav & Wind Speed (m/s) & 22.95& 35.03 & 27.39 & 1.67 \\ 
\hline
TQL\_dav & Liquid precipitation (liters/m\textsuperscript{2}) & 0 & 0.477& 0.144 & 0.087 \\ 
\hline
\end{tabular}
\end{table}

\subsection{Forecasting Algorithm: XGBoost}
XGBoost is an extension of gradient boosting, a powerful ensemble technique for both classification and regression tasks. Gradient boosting combines weak base learners (trees) into a strong learner, with each learner correcting the errors of its predecessors. XGBoost improves upon traditional gradient boosting by incorporating advanced regularization techniques to prevent overfitting and enhance generalization. It also leverages second-order gradients of loss functions and has the inherent ability to handle missing values [20]. Previous research has demonstrated its competence in various time series forecasting tasks, such as sales forecasting [21] and pandemic prediction [22].We chose XGBoost as our surrogate forecasting algorithm primarily due to its shorter runtimes, making it particularly well-suited for evaluating the effectiveness of HPO techniques, especially when compared to computationally intensive neural network models [23]. Table 3 displays the key XGBoost hyperparameters in our analysis, along with their short descriptions and the experimented values.

\begin{table}
\centering
\caption{Experimented Key XGBoost Parameters}
\label{tab:xgboost_params}
\begin{tabular}{|p{1.8cm}|p{2.2cm}|p{2.2cm}|} 
\hline
\textbf{Parameter} & \textbf{Description} & \textbf{Experimented Values} \\
\hline
`max\_depth` & Maximum depth of a tree & 3, 4, 5, 6, 7, 8, 9, 10 \\
\hline
`learning\_rate` & Step size shrinkage used in update to prevents overfitting & 0.001, 0.003, 0.005, 0.007, 0.009, 0.01, 0.03, 0.05, 0.07, 0.09, 0.1, 0.3, 0.5, 0.7, 0.9 \\
\hline
`n\_estimators` & Number of trees in the ensemble & 100, 200, 300, 400, 500, 600, 700, 800, 900, 1000 \\
\hline
`subsample` & Subsample ratio of the training instances & 0.5, 0.7, 0.8, 1.0 \\
\hline
`colsample\_bytree` & Subsample ratio of columns when constructing each tree & 0.5, 0.6, 0.7, 0.8, 0.9, 1.0 \\
\hline
`min\_child\_weight` & Minimum sum of instance weight (hessian) needed in a child & 1, 3, 5, 7 \\
\hline
\end{tabular}
\end{table}

\subsection{Hyperparameter Optimization (HPO) algorithms}

To systematically explore the hyperparameter space and identify optimal configurations for XGBoost in STLF tasks, we employed five HPO algorithms:
\begin{itemize}
\item Random Search:A simple technique that randomly samples hyperparameter values from a predefined distribution for a specified number of iterations. We chose this method as a representative of 'uninformed' HPO methods, which lack the ability to conduct a guided search in hyperparameter space. 
\item CMA-ES: A powerful evolutionary optimization algorithm that samples candidate solutions from a multivariate normal distribution and adapts the distribution's mean and covariance matrix over generations to guide the search towards better solutions. This derivative-free and stochastic approach effectively avoids getting stuck in local optima, while the covariance matrix adaptation learns the underlying structure of the search space to improve efficiency [24].
\item Bayesian Optimization: A sequential model-based optimization algorithm that incorporates prior beliefs about the objective function, typically represented by a Gaussian Process. It intelligently selects the next hyperparameter configuration to evaluate using an acquisition function, balancing the exploration of new areas of the parameter space with the exploitation of promising regions. After each evaluation, the result is used to update the prior belief, refining the model's understanding of the objective function. Key features include sample efficiency (requiring fewer evaluations than grid or random search), the ability to handle black-box functions (where the underlying relationship between hyperparameters and model performance is unknown), and the ability to quantify uncertainty in predictions.
\item Partial Swarm Optimization (PSO): A computational method that optimizes solutions by simulating the social behavior of a swarm (e.g., birds, fish). Each particle in the swarm represents a potential solution, and its movement is influenced by its own best-known position and the best-known position of the entire swarm [25]. Key features include its simplicity, derivative-free nature, and ability to explore the global search space.
\item Nevergrad Optimizer (NGOPT): A meta-algorithm within the Nevergrad library that automatically selects and combines different optimization strategies based on the problem's characteristics, aiming to efficiently find good solutions [26]. NGOpt's adaptability distinguishes it from other HPO algorithms, as it dynamically adjusts its approach to the specific optimization landscape. To the best of our knowledge, this study is the first to explore the application of NGOpt for optimizing XGBoost hyperparameters.
\end{itemize}

\subsection{Experiments }
We run experiments on the Google Colab Pro+ platform with an A100 GPU.The python implementation uses hyperopt for bayesian optimization, Neverggrad for CMA-ES, PSO, and NGOpt, and sckitlearn for the random search. The experiments involve various model configurations. A model has the following components: a dataset, a univariate or multivariate sample, an XGBoost Algorithm for STLF, and an HPO algorithm. The selected HOP algorithms search the XGBoost hyperparameters displayed above in Table 2  in 50 experiments. We set the early stopping at 20 for the selected HPO algorithms, except for random search. Random search uses a 5-fold cross-validation strategy for hyperparameter search. RMSE was the metric chosen to be minimized for the optimization process.

\subsection{Performance Evaluation }
We employed three metrics for performance evaluation: Mean Absolute Percentage Error (MAPE), R-squared (R2), and runtime. Importantly, the runtime was measured as the total time taken by each HPO algorithm to find the optimal hyperparameters for XGBoost, starting from the initialization of the objective function. These metrics were recorded for each model configuration. Subsequently, we generated performance plots to visualize the evolution of these metrics across sample sizes ranging from 1,000 to 20,000 observations, with increments of 1,000. These plots provide valuable insights into the scalability and performance trends of the HPO algorithms under different data volumes.  It is important to note that the values displayed in the performance plots are scaled to the 0-1 interval for better visualization and comparison across metrics.

\subsection{Testing Between-group Differences}

Following analyzing the performance plots, we investigated the statistical significance of differences between the HPO algorithms' performance metrics at each sample size. For this purpose, we conducted a non-parametric Kruskal-Wallis test (a rank-based alternative to one-way ANOVA) followed by pairwise comparisons with Bonferroni correction to control the familywise error rate at a significance level of $\alpha = 0.05$. The following is the null hypothesis ($H_0$):
\begin{equation}
H_0: MR_1 = \dots = MR_k
\end{equation}
The test results are reported in a matrix format, displaying the mean rank (MR) differences between algorithm pairs along with their associated p-values. The mean rank (MR) is calculated by dividing the sum of ranks within a group by the number of observations in that group.

\section{RESULTS AND DISCUSSION}
Our main results demonstrate the superiority of sequential model-based optimization (SMBO) and population-based methods over random search. The following Kruskal-Wallis test revealed significant differences in runtime performance at  $\alpha =0.05$, with SMBO and population-based methods exhibiting significantly shorter runtimes. This can be attributed to their distinctive features, such as guided exploration of the hyperparameter space, information sharing among candidate solutions, and early stopping mechanisms that prevent unnecessary iterations [27]. For univariate forecasting models, Bayesian optimization yielded the lowest accuracy across the selected metrics, although its performance was not statistically different from NGOpt except for R2. Random search was the slowest algorithm in identifying optimal hyperparameters, while CMA-ES was significantly slower than PSO.

\begin{table*}[!ht]
\centering
\caption{Comparison of Hyperparameter Optimization (HPO) algorithms on MAPE, R\textsuperscript{2}, and Runtime metrics.}
\label{tab:hpo_comparison}
\begin{tabular} {llllllllll}
\hline
\multicolumn{2}{c}{} & \multicolumn{4}{c}{\textbf{Univariate}} & \multicolumn{4}{c}{\textbf{Multivariate}} \\
\hline
\multicolumn{2}{c}{} & {Min} & {Max} & {Mean} & {sd} & {Min} & {Max} & {Mean} & {sd} \\
\hline
\textbf{MAPE} 
    & Random Search & 0.110
 & 0.124 & 0.115 &0.003  & 0.08 & 0.117 & 0.092  & 0.012  \\
    & CMA-ES &0.110 &0.125  &0.115  &0.004  &0.084  &0.124  & 0.097 & 0.013 \\
    & Bayesian &0.115  & 0.125 & 0.115 & 0.004 & 0.083	
 & 0.120 & 	0.095	 & 0.011 \\
    & PSO & 0.110	
 & 0.129	 & 0.116 & 	0.005 & 0.081	
 & 0.124	 & 0.096	 &  0.013\\
    & NGOpt & 0.110
 & 	0.129	 & 0.116	 & 0.005 & 0.083	
 & 0.123	 & 0.096	 &  0.012\\
\hline
\textbf{R\textsuperscript{2}}
    & Random Search & 0.243	
 & 0.340	 & 0.293	 & 0.029 & 0.318	

 & 0.576 & 	0.497 &  	0.054\\
    & CMA-ES & 0.222	
 & 0.338 & 	0.292	 & 0.030 & 0.272	
 & 0.545	 &0.463 & 	0.067  \\
    & Bayesian & -0.356	
 & 0.293	 & 0.093 &0.187 &0.146	
  & 0.538 & 	0.473	 &  0.083\\
    & PSO &0.180	
  & 0.332& 	0.268	 &0.037   & 0.238	
 & 0.541	&0.461	  & 0.080  \\
    & NGOpt &0.165	
  &0.3232  & 	0.261	 & 0.04 &0.198	
  & 0.524	& 0.462	  &  0.073\\
\hline
\textbf{Runtime}
    & Random Search &69.908	&109.358& 93.683& 	10,828 &135.662	
	&379.940		& 298.492	 & 69.895  \\
    & CMA-ES & 2.511	
 & 9.434	 &4.409	 & 1.760  &4.538	
  &42.017	  & 16.277	 &8.526  \\
    & Bayesian &2.590
  & 	11.399	 & 6.277& 2.376 &6.839	
  &33.359	  &19.713	  & 7.753 \\
    & PSO &2.891	
  &12.884  &	7.377&2.299  &9.171
  &23.713  &	17.385	  &3.668  \\
    & NGOpt &2.245	
  &20.163	  &7.635&4.277   &5.073	
  &26.218	&13.464&5.204   \\
\hline
\end{tabular}
\end{table*}

In multivariate analysis, all HPO algorithms improved XGBoost performance with increasing sample size, as evidenced by the decreasing MAPE trend in the scalability plots. Notably, the statistical differences between accuracy performances became insignificant, suggesting that Bayesian HPO can effectively leverage contextual features like weather and calendar information to enhance its performance in this setting. This aligns with the findings of [27], who demonstrated that incorporating relevant features can improve the accuracy of Bayesian optimization for time series forecasting tasks. Similar to the univariate analysis, random search remained the slowest algorithm.

\begin{table*}[!ht]
\centering
\caption{The Kruskal Wallis test results}
\label{tab:kruskal_wallis}
\begin{tabular}{llllllllll}
\hline
\multicolumn{2}{c}{} & \multicolumn{4}{c}{\textbf{Univariate}} & \multicolumn{4}{c}{\textbf{Multivariate}} \\
\hline
\multicolumn{2}{l}{}                       & CMA-ES & Bayesian & PSO    & NGOpt  & CMA-ES & Bayesian & PSO   & NGOpt  \\
\hline
\textbf{MAPE}   & Random   Search & 2.4    &\textbf{ -43.2*}     & -0.6   & -2.35  & -17.05 & -11.05   & -14.7 & -15.95 \\
                         & CMA-ES          & 0      & \textbf{ -45.6*}    & -3     & -4.75  & 0      & 6 & 2.35  & 1.1    \\
                         & Bayesian        & \textbf{ -45.6*} & 0        & \textbf{ 42.6*}   & 40.85  & 6      & 0        & -3.65 & -4.9   \\
                         & PSO             & -3    & \textbf{ 42.6*} &0& 1.75  & 2.35   & -3.65    & 0     & -1.25  \\
                         \hline
\textbf{R\textsuperscript{2}}     & Random   Search & 0.8   & \textbf{47.9*}     & 18.2  & 19.9  & 21.8  & 12.4     & 17.4 & 20.2  \\
                         & CMA-ES          & 0      & \textbf{47.2*}    & 17.4   & 19.2   & 0      & -9.4     & -4.5 & -1.6   \\
                         & Bayesian        & \textbf{47.2*}  & 0        & \textbf{-29.8*} & \textbf{-27.2*} & 9,4    & 0        & 4.9  & 7.8    \\
                         & PSO             & 17.4   & \textbf{-29.8*}   & 0      & 1.8    & -4.5  & 4.9     & 0     & 2.9   \\
                         \hline
\textbf{Runtime} & Random   Search & \textbf{66.8*}  & \textbf{49.4*}    & \textbf{39.5*}   & \textbf{44.4*}   & \textbf{53.9*}   & \textbf{41.1*}    & \textbf{44*}    & \textbf{61*}    \\
                         & CMA-ES          & 0      & -17.4    & \textbf{-27.3*}& -22.4 & 0      & -12.9   & -9.9  & 7.2   \\
                         & Bayesian        & -17.4  & 0        & -9.9  & -4.9  & -12.9 & 0        & 2.9  & 20     \\
                         & PSO             & -27.3 & -9.9    & 0      & 4.9    & -9.9   & 2.9     & 0     & 17.1 \\
                         \hline
\multicolumn{10}{l}{\footnotesize * Statistically significant at $\alpha = 0.05$} 
\end{tabular}
\end{table*}

\begin{figure*}[!ht]
     \centering
     \subfloat[]{\includegraphics[width=0.32\textwidth]{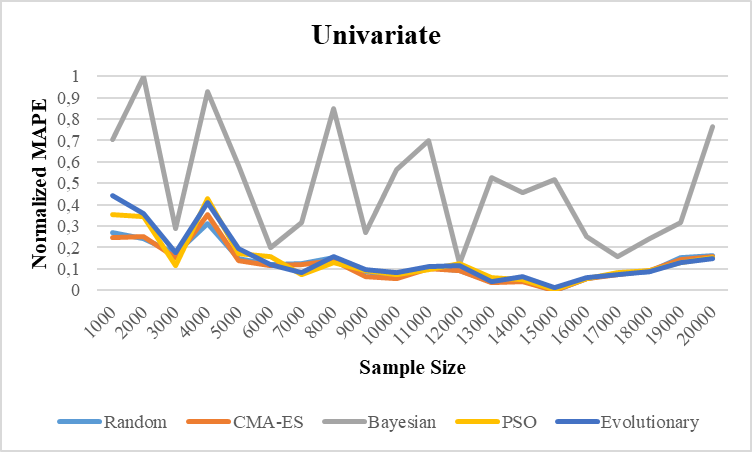}}          \subfloat[]{\includegraphics[width=0.32\textwidth]{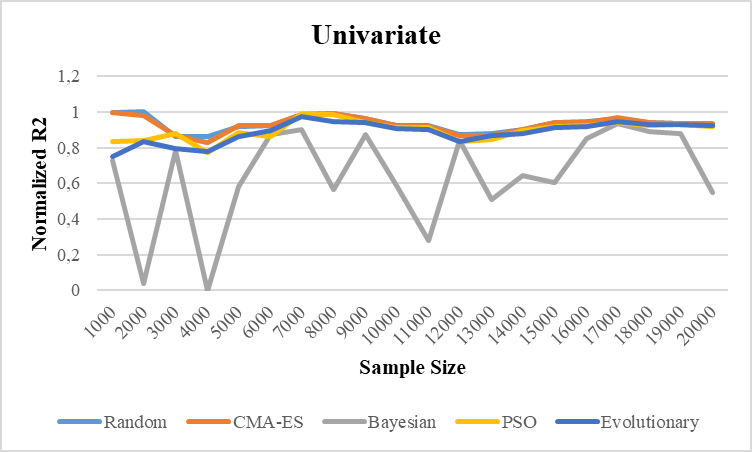}} 
     \subfloat[]{\includegraphics[width=0.32\textwidth]{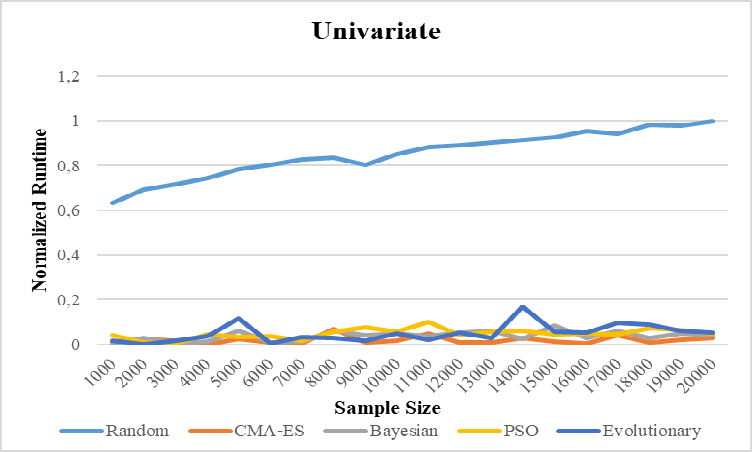}}
     
     \subfloat[]{\includegraphics[width=0.32\textwidth]{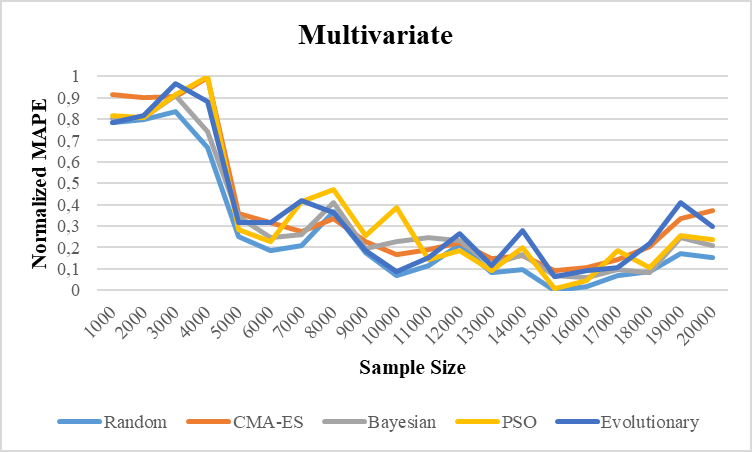}}  \subfloat[]{\includegraphics[width=0.32\textwidth]{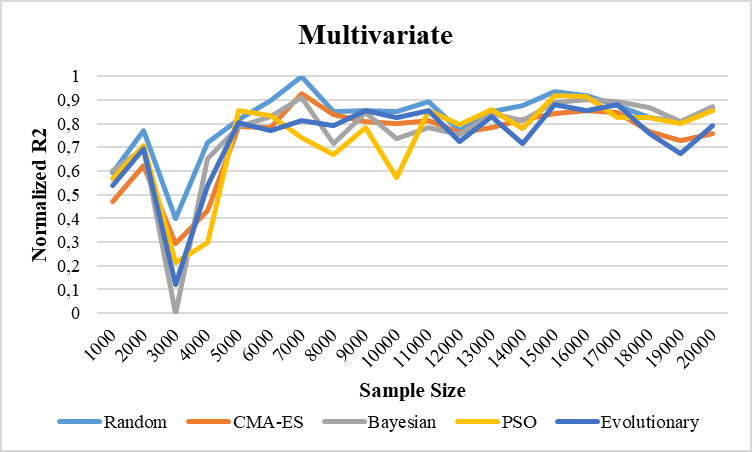}}  \subfloat[]{\includegraphics[width=0.32\textwidth]{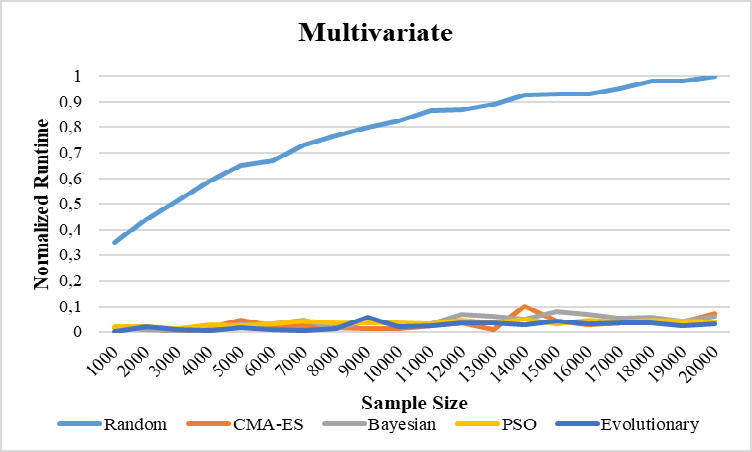}}
     \caption{Performance Plots of HPO algorithms in  MAPE across (a) univariate (d) multivariate models, in R\textsuperscript{2} across (b) univariate  (e) multivariate models, and in runtime across (c) univariate  (f) multivariate models}
     \label{fig:word_char_dist_doors}
 \end{figure*}

\section{CONCLUSION}
This study has examined the efficacy of five HPO algorithms in the univariate and multivariate STLF context. These algorithms were tested on a well-known surrogate time series forecasting algorithm, XGboost, based on their MAPE, $R^2$, and runtimes. The performance of the HPO techniques is displayed on the scalability plots that show the selected metrics across sample sizes between 1000 and 20000 with a 1000 increase. The results have showcased the effectiveness of the selected intelligent HPO techniques against the baseline, random search, particularly regarding runtimes. Another result indicates the incapability of Bayesian optimization in the univariate settings.

This study has several limitations. First and foremost, the computational complexity and associated longer runtimes restrained the analysis only to a single forecasting algorithm (XGBoost).  Second, the same reasons severely limit the number of the experimented hyperparameters and their values. For example, minimum split loss and L2 regularization terms on weights can be incorporated into the analysis in future research. Third, similar to the previous limitations, we experimented with only five HPO algorithms. 
Also, the selected techniques have a host of different search algorithms; these can be incorporated into the experiments as well. Lastly, the efficacy of the  HPO techniques was tested in only two datasets, which limits its generalizability to other contexts.

Several avenues for future research emerge from this study. First, incorporating interpretability metrics could provide deeper insights into the influence of specific features on model performance. Second, a comparative analysis of model performance across different countries or regions could reveal geographic variations and inform the development of tailored forecasting strategies. In a similar vein, comparing performance across building types (e.g., residential, commercial) could further assess the robustness of HPO algorithms and identify potential domain-specific optimizations. Finally, investigating the scalability of HPO algorithms across different time intervals in STLF (e.g., hourly, daily, weekly) could uncover temporal patterns and improve forecasting accuracy at various granularities.

\vspace{12pt}

\end{document}